\documentclass[twocolumn, 10pt]{article}
\usepackage[a4paper, margin=1in]{geometry} 
\usepackage{graphicx} 
\usepackage{amsmath, amssymb} 
\usepackage{hyperref} 
\usepackage{titlesec} 
\usepackage{caption} 
\usepackage{float} 
\usepackage{booktabs}
\usepackage{array}
\usepackage[numbers]{natbib}
\usepackage{tabularx}

\title{\textbf{Integrating emotional intelligence, memory architecture, and gestures to achieve empathetic humanoid robot interaction in an educational setting }}
\author{Fuze Sun, Lingyu Li, Shixiangyue Meng, Xiaoming Teng, Terry R. Payne, Paul Craig \\ \textit{University of Liverpool \& Xi'an Jiaotong-Liverpool University}}
\date{\today}

\titleformat{\section}{\large\bfseries}{\thesection.}{1em}{}
\titleformat{\subsection}{\normalsize\bfseries}{\thesubsection.}{1em}{}
\begin{document}

\twocolumn[
\maketitle

\begin{abstract}
This study investigates the integration of individual human traits into an empathetically adaptive educational robot tutor system designed to improve student engagement and learning outcomes with corresponding Engagement Vector measurements. While prior research in the field of Human-Robot Interaction (HRI) has examined the integration of the traits, such as emotional intelligence, memory-driven personalization, and non-verbal communication, by themselves, they have thus-far neglected to consider their synchronized integration into a cohesive, operational education framework. To address this gap, we customize a Multi-Modal Large Language Model (LLaMa 3.2 from Meta) deployed with modules for human-like traits (emotion, memory and gestures) into an AI-Agent framework. This constitutes the robot’s intelligent core that mimics the human emotional system, memory architecture and gesture controller to allow the robot to behave more empathetically while recognizing and responding appropriately to the student’s emotional state. It can also recall the student’s past learning record and adapt its style of interaction accordingly. This allows the robot tutor to react to the student in a more sympathetic manner by delivering personalized verbal feedback synchronized with relevant gestures. Our study suggests the extent of this effect through the introduction of Engagement Vector Model which can be a benchmark for judging the quality of HRI experience. Quantitative and qualitative results demonstrate that such an empathetic responsive approach significantly improves student engagement and learning outcomes compared with a baseline humanoid robot without these human-like traits. This indicates that robot tutors with empathetic capabilities can create a more supportive, interactive learning experience that ultimately leads to better outcomes for the student.  
\end{abstract}
\vspace{0.5cm}
]

\section{Introduction}
With advancements in AI technologies, the capability of robots to interact with humans in a more personalized and intuitive manner has evolved at a rapid pace \cite{obaigbena2024ai}. In educational settings, empathetic strategies—such as emotionally intelligent responses and cognitive learning techniques have been shown to enhance student engagement and learning outcomes  \cite{bozkurt2010relationship}. These findings have motivated researchers to implement empathetic robot technologies in education for promoting student’s learning performance based on customized learning strategies and activities \cite{chu2022artificial}.  However, there are still major challenges to be overcome before robots can be considered to accurately detect, interpret, and respond to student’s emotional and cognitive states in real-time \cite{dautenhahn2007socially}.

Real-time natural human-like communication is challenging because of the implicit depth and subtlety of human behavior. Patterns of human behavior have evolved over millennia and are deeply rooted in cognitive processes, emotional states, and learned behavior. Unlike human teachers, most robot systems lack the ability to coordinate key human-like traits, such as emotional intelligence, memory-based adaptation, and gestural communication. This limits their effectiveness in creating meaningful interactions where emotions and empathy play a critical role in instructional responses and learning motivation \cite{meyer2002discovering}. A robot that cannot demonstrate this kind of genuine behavior risks being perceived as rigid and impersonal, limiting its ability to foster an engaging classroom environment.  

To address these challenges, this study proposes an AI-driven robotic tutor that integrates a Multi-Modal AI agent as its intelligent core, combining three essential components for human-robot interaction (HRI). These are as follows.  

\begin{itemize}
    \item \textbf{Emotional Intelligence} – Enables the robot to analyze student emotions (e.g., frustration, interest, excitement) and reply accordingly. 
    \item \textbf{Memory Architecture} – Allows the robot to retain and recall student learning histories and preferences, personalizing interactions based on prior engagement. 
    \item \textbf{Gesture Control System} – Enhances non-verbal communication, linking verbal instructions to physical gestures (e.g., pointing, tracing) to ground abstract concepts to a tangible avatar. 
\end{itemize}

By integrating these elements, we can envision that a robot tutor would have the capability to move away from more rigid, pre-programmed responses, instead generating real-time, adaptive interactions with a more realistic emotional response that has greater potential to form a positive impact on social and mental well-being \cite{molero2020emotional}. The generated related speech content, along with appropriate gestures that ground the teacher’s speech by linking abstract verbal cues to the tangible, physical environment—such as pointing or tracing—can help achieve human-like grounding behavior during the teaching process \cite{valenzeno2003teachers}. By merging these capabilities, the robot tutor can dynamically adapt to each student's engagement level, offer tailored explanations, and provide positive reinforcement and encouragement to mitigate negative emotions such as anxiety, which has been shown to reduce working memory capacity and impair the student's learning experience \cite{paas2014cognitive}. Together, these traits foster a more empathetic and responsive learning environment.

\begin{figure}[H]
    \centering
    \includegraphics[width=\linewidth]{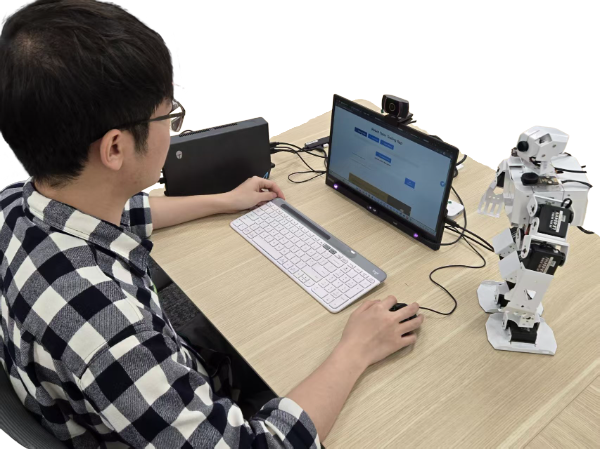}
    \caption{Experimental setup showing a student interacting with an AI-powered robot tutor during a history lesson. Technical details can be found in Figure 3.}
    \label{fig:robot_home}
\end{figure}

This research applies an AI-powered robot tutor in an educational scenario (see Figure~\ref{fig:robot_home}), using history lessons—such as the \textit{Apology of Socrates}—as a case study. Leveraging its multimodal AI capabilities, the robot recognizes engagement cues, dynamically adjusts its teaching style, and builds rapport with learners through personalized interactions. The AI agent’s knowledge base supports long-term adaptation to each student’s learning progress, ensuring that past learning records and personal preferences inform future interaction strategies. This study explores how robots can better understand and respond to students’ behaviors, emotions, and social cues to maintain engagement over time (i.e., long-term human-robot interaction), thereby enhancing learning outcomes in educational settings.

The main objectives of this research are to firstly develop a Multi-Modal AI Agent based robot tutor capable of empathetic interaction, and evaluate its impact on student engagement, learning retention, and satisfaction, compared to a robot without these empathetic traits. Finally we want to analyze the broader potential of empathic intelligent robots in fostering interactive and supportive learning environments. 

While previous research in the field of Human-Robot Interaction (HRI) has largely investigated individual human-like traits—such as emotion, memory, or gestural communication—in isolation to improve interaction quality in educational settings, a significant gap remains in understanding how these characteristics can be integrated to create a holistic, real-world empathetic interaction experience that combines multiple aspects of human behavior. Addressing this gap, our research uniquely integrates emotional adaptability, memory-driven personalization, and non-verbal communication into a unified system, resulting in an AI-powered robot tutor capable of delivering a customized and engaging learning experience. To achieve this, we have designed and developed an application that integrates multiple empathetic traits into a cohesive system, enabling a level of interactive emotional engagement and personalization not previously demonstrated in educational robotics research. Through this work, we aim to advance the field by illustrating how the integration of multiple human-like capabilities within a robot tutor can significantly enhance student engagement, motivation, and learning outcomes.

\section{Related Work}
Human-Robot Interaction (HRI) has become an increasingly prominent field of research, with robotic agents being deployed across a wide range of domains, including education, healthcare, entertainment, and customer service \cite{meyerson2023introductory}. As robots engage more directly with humans, it is essential that they are capable of understanding and responding to human emotions, needs, and preferences in an empathetic manner. This review of related work provides a comprehensive overview of the current state of the art in four critical areas that contribute to effective HRI. By critically examining the existing body of research in these areas, the review identifies key trends, persistent challenges, and research gaps, highlighting opportunities to advance the development of more empathetic, adaptive, and engaging robot tutors—particularly in educational contexts where personalized and empathetic interaction is crucial for fostering student engagement and improving learning outcomes.

\subsection{Emotional Intelligence and Empathic Responsiveness in HRI}
Emotion, which can contribute to achieving empathy, is a critical factor in human-robot interaction, especially in educational contexts where understanding and emotional connection can enhance the learning experience and subsequent outcomes as emotions can arise in response to the meaning structures of the current situations, and how an individual interprets these scenarios act as the catalyst for different experiences \cite{schutz2006reflections}. Research has revealed that robots equipped with emotional models such as OCC (Ortony, Clore, and Collins) can recognize, interpret, and adapt to human emotions. Therefore, they tend to foster greater engagement, trust, and satisfaction among users \cite{kwon2007emotion}. Facial expressions are important cues to communicate one’s emotions or intentions to others , and robots need to be able to understand human expressions to predict emotions and appropriately respond to them \cite{rawal2022facial}. The robot first infers the emotional state of the user and then accordingly generates facial expressions and gestures in response to their audience. \emph{Parallel Empathy} (i.e. the same emotion as the audience) or \emph{Reactive Empathy} (where the robot responds to the audience’s emotion) strategies both contribute to the generated emotions nested in the interaction process. In the area of Human-Robot Interaction (HRI), empathy is defined as the robot’s ability and process to recognize the user’s emotional state, thoughts and situation, and produce affective or cognitive response to make humans experience positive perception \cite{park2022empathy}. 

Among current emerging technologies, robots are believed to have the significant potential to shape the future of education and lead students to better learning outcomes. Related learning experiences from educational robots such as coding, or collaborating with, can, for example, inspire creativity, train coding problem and empower self-regulated learning of students, which can help to shape them to be successful in today’s and tomorrow’s working environment \cite{alnajjar2021robots}. In educational settings, robots as social agents can take over some of the tasks which humans have traditionally had to perform, such as providing academic guidance. However, due to the limitation of intelligence, robots now on the market become the learning companions instead of the teacher which is an essential part of the digital turn in future education. Robots with emotional intelligence have been found to improve student motivation and learning by responding with warmth and understanding. For example, studies on social robots show that students respond more positively when robots use verbal cues, such as offering encouragement or other forms of positive reinforcement when students appear frustrated or confused \cite{brown2014positive}. 

Nevertheless, most existing systems treat emotional intelligence as a standalone capability, relying heavily on pre-defined emotional models and fixed response strategies that limit adaptability across diverse contexts. These systems normally focus on immediate, reactive empathy showcasing rather than a sustained emotional interaction that accounts for evolving student's needs and long-term interaction data. Given the growing importance of emotional intelligence in HRI and its potential benefits for a student’s learning process, it is necessary to explore innovative approaches to coordinate emotional understanding with other essential human-like traits into robot interactions with students. 

\subsection{Memory-Driven Personalization for Adaptive Human-Robot Interaction}
The memory system within robots enables personalized interaction by allowing the robot to retrieve previous knowledge of individual users, thus facilitating customized conversation \cite{ligthart2022memory}. In educational scenarios, memory-driven personalization can improve learning outcomes by adapting instruction based on a student’s prior knowledge and preferred learning style to show empathy. What is more, personalized education can increase situational interest of learning in the short term, which is a necessary factor for enhancing an individual’s interest in the long-term from a psychological view \cite{reber2018personalized}. Episodic memory, which allows robots to store and retrieve previously acquired information, is particularly effective in building rapport, as users feel recognized and valued when the robot recalls  their past interactions with them \cite{leyzberg2018effect}.  This past interaction record, in fact, is a personal profile of the user stored by the robot system to give the impression of familiarity. Memory architectures in HRI research address the value of such approaches in fostering long-term relationships between humans and robots \cite{lee2012personalization}, which is essential in educational settings where personalization and rapport are important for student engagement. 

Recent advancements in integrating memory functions into large language models provide new opportunities for personalization. The \emph{LangChain} framework aims to build applications powered by LLMS, and therefore offers reliable tools for incorporating memory into interactions \cite{topsakal2023creating}. LangChain enables the creation of both short-term memory (details of current interaction) and long-term memory (storing and recalling past interactions). Another emerging framework, \emph{Ollama}, supports the creation of a knowledge base where info documents can be stored and retrieved for use with LLMs \cite{marcondes2025using}. In this scenario, we use Ollama with Llama 3.2 from Meta to construct our AI-Agent for the experiment. This allows for setting up a local repository of data, such as PDF or DOC documents for user information. By leveraging large language models with memory functions in the experiment, the study aims to take a significant step forward in developing robots that can not only understand but also remember individual student’s needs and preferences. This personalization approach is expected to further enhance engagement, motivation, and learning outcomes through offering tailored support and fostering deeper relationships between the robot tutor and students. 
\subsection{Non-Verbal Communication through Gestures and Social Cues in HRI}
Non-verbal communication such as gestures, improve human-robot interaction by making the interaction more natural and intuitive.  Research has shown that eye contact helps establish a sense of attentiveness in human-robot interaction \cite{admoni2017social}, which is crucial for creating a supportive educational environment and address the challenges of managing turn-taking during conversation. Studies have also found that non-verbal cues are important for maintaining engagement and interest, since conversation partners interpret these cues as showing empathy and understanding \cite{rodriguez2015bellboy}. What is more, the robot can utilize these social cues to affect the perceptions of the mental states and intentions of the user to achieve bidirectional communication \cite{fiore2013toward}. 

Gestures play an essential role in emphasizing and clarifying verbal communication whilst conveying empathy; for example, a robot using hand gestures to illustrate the size or importance of historical events and enhance understanding or engagement. In addition, when the student is struggling with a difficult concept, the robot’s tone could be softer, and it could nod or blink to mimic attentiveness and concern. People who participate in HRI tend to direct their attention to the robot more often in interactions where engagement gestures are implemented, and feedback reveals that interactions are more appropriate compared to not having engagements \cite{sidner2005explorations}.  In educational settings, non-verbal social cues, such as eye gaze, index finger hand shape or leaning forward, can express inner emotion that supports the building of rapport, reinforcing the robot’s role as a supportive teaching assistant in the educational setting. Combining social behavior and social cues congruently together contributes to maximizing the learning outcome in HRI \cite{kennedy2017impact}.  

While numerous studies affirm the advantages of non-verbal communication, for example, gestures and eye gaze in fostering HRI experience, most existing models adopt limited or pre-scripted behaviors that lack contextual adaptability. For instance, although hand gestures can clarify abstract concepts and aid comprehension in educational settings, many robotic systems deploy gestural modules in a rigid, predefined manner without real-time synchronization with speech or student feedback. This one-size-fits-all approach undermines the robot's ability to respond meaningfully to student needs in dynamic learning environments. To truly utilize the power of non-verbal cues in communication, future educational systems must go beyond isolated gesture modules and integrate them into Multi-Modal AI-Agent frameworks that dynamically adapt to students' behaviour, states, and references

\subsection{Integrating Multi-Modal Language Models for Context-Aware Robotic Interaction }
Multi-Modal large language models such as LLaMA, represent a breakthrough in edge AI and vision with open, customizable models. Supported by a broad ecosystem, Llama 3.2 11b is a direct replacement for the previous unimodal text models, while being greatly enhanced on image understanding tasks compared to closed models, such as Claude 3 Haiku. Both pre-trained and aligned models included in Llama 3.2 are available that can be fine-tuned for custom applications, and deployed locally compared to other multimodal models \cite{meta2024llama}. This kind of model enables robots to engage in complex, contextually relevant conversations by understanding input modes such as text, audio, and visual cues. Multi-Modal LLMs leverage extensive training data to enable comprehensive understanding, generate reasonable response, and adapt gestures across different types of interactions. Research reveals that these advantages are essential for robots implemented in educational settings, where the robot as the tool has the potential to foster engagement and motivation  \cite{tozadore2024teachers}. 

\begin{figure}[H]
    \centering
    \includegraphics[width=\linewidth]{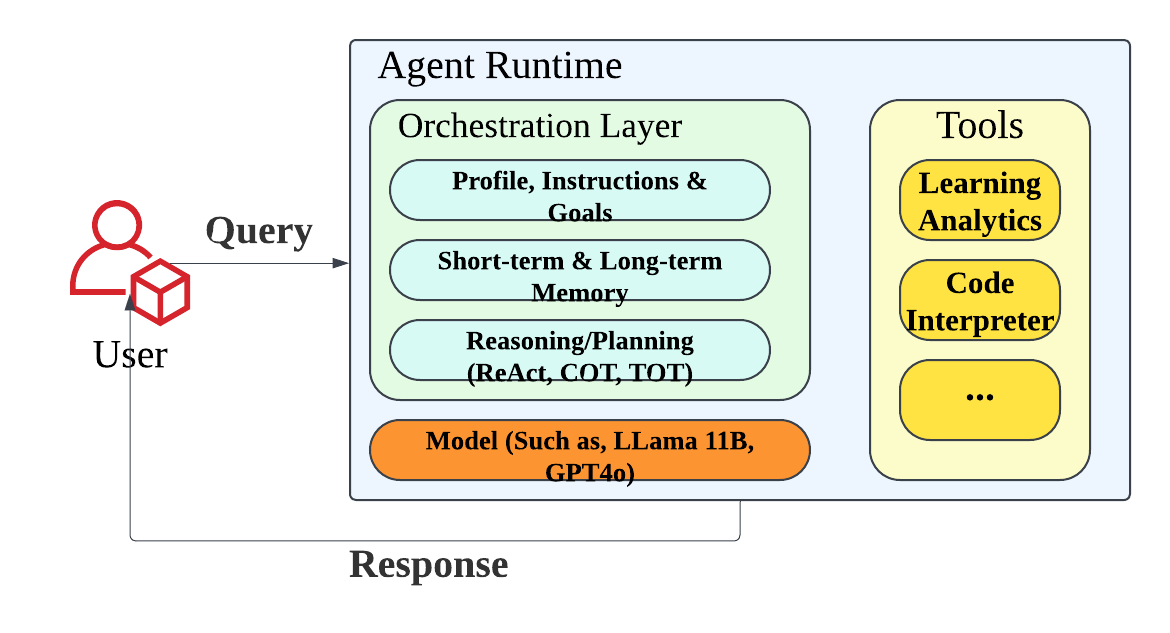}
    \caption{Layered architecture of the AI Agent Framework used as the robot’s cognitive core. The system integrates a reasoning-capable orchestration layer, memory mechanisms, and LLMs (e.g., GPT-4o, LLaMA) with specialized tools such as learning analytics and code interpreters. }
    \label{fig:setup}
\end{figure}

The LLM has the potential to reshape teaching from an educator's perspective by equipping them with a wide range of tools and resources which can assist with lesson navigation, personalization content development, differentiation instructions based on the student’s need, and quiz assessment. It can also be used to automatically generate questions, explanations, and assessments that are tailored to the student’s level of knowledge for them to learn at their own pace in tutoring scenarios \cite{kasneci2023chatgpt}. The personalized interventions provided by LLMs in education, which use a student’s individual values, and preferences can help to increase interest in school subjects \cite{meyer2023chatgpt}. Therefore, this research integrates these teaching functionalities into the AI-Agent to exhibit behaviors similar to those of a responsible tutor. What is more, by utilizing the Multi-Modal LLM’s advantages in a layered architecture that mimics the emotional system, memory architecture, and gestures control as a human-being, robots gain the functions needed to recognize student engagement levels, recall individual preferences, and adjust their emotional tone (see Figure~\ref{fig:setup}). This adaptability makes the AI Agent Framework particularly valuable for robots with empathy in education, as they can continuously refine their expression, gesture, and verbal communication based on the interaction context and individual learning needs, forming a highly interactive and personalized experience for students. 

Building upon the findings from these related studies, we aim to investigate whether an empathetic robot system equipped with multiple coordinative human-like traits can work together to provide more tailored support to students than previously developed robots which demonstrate basic affective behaviors, but often lack the capacity to coordinate emotional understanding with other critical human-like traits. This narrow focus reduces the depth and realism of interactions, making it difficult for students to form meaningful connections with robot tutors. As a result, many current approaches fall short in fostering long-term engagement or adapting to individual learning trajectories.

For now, there is a growing need for integrative approaches that coordinate different characteristic modules under a broader Multi-Modal framework. We customize a Multi-Modal LLM into an educational AI-Agent Framework with surrounding human-trait modules for controlling the robot to enable personalized and adaptive interactions while synchronizing them with verbal and non-verbal cues. The research hypothesizes that the embodied robot system powered by that AI-Agent will not only demonstrate higher levels of empathy but also lead to better learning outcomes for students. In the next section, the details of the experimental study will be presented, which compares the performance of the intelligent robot system to the robot without or with few empathetic elements in terms of student engagement, motivation and learning outcomes. 

\section{Robot System: Functional Approach and Configuration}

\begin{figure*}[t]
    \centering
    \includegraphics[width=\textwidth]{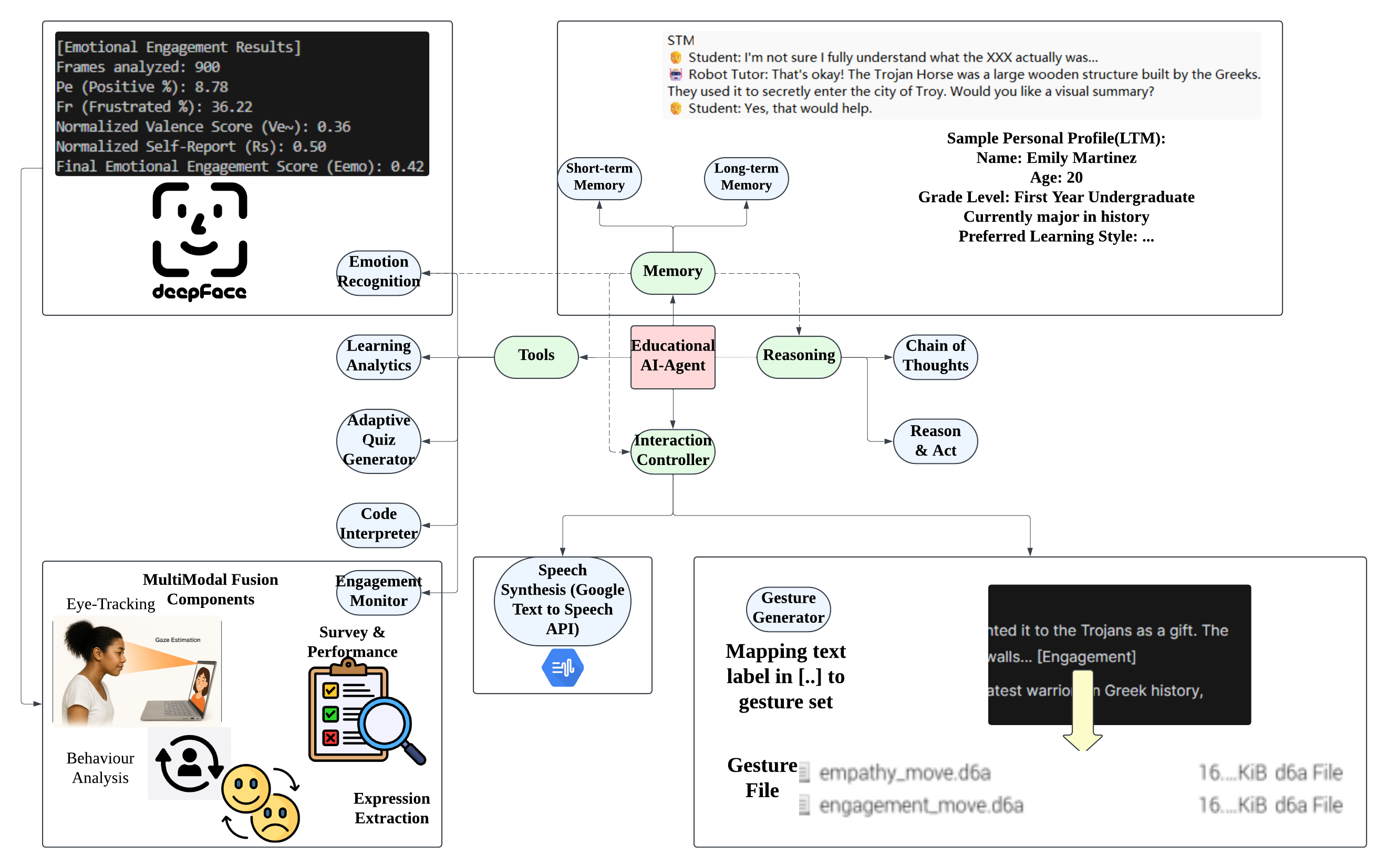}
    \caption{Architecture of the AI-Agent powering the robot tutor, integrating multimodal input processing, memory-based reasoning, and gesture-verbal coordination for empathetic interaction.}
    \label{fig:Agent}
\end{figure*}

The robot tutoring system is designed to simulate a history learning experience that emphasizes interaction and adaptive teaching behavior. A laptop, used to display course materials and record quiz answers, is placed on an adjustable stand at eye level. The robot agent is positioned to the right of the table, between the laptop and the student, allowing the student to easily see and interact with both the robot and the laptop. The technical setup consists of three main components. These are the robot agent based on a Raspberry Pi, the Course App running on the laptop, and the laptop hosting the Ollama server. Both devices are connected to the same Wi-Fi network, enabling seamless communication between them. The robot platform, built on a Raspberry Pi, includes a camera with facial recognition functionality, a microphone for capturing student speech and a gesture management system. The laptop is responsible for running both the Course App (see Appendix~\ref{appendix:courseapp}) and the AI agent (see Figure~\ref{fig:Agent}), which serves as the cognitive engine of the robot tutor. The internal components of the robot will be detailed below to support future replication.

\begin{figure}[H]
    \centering
    \includegraphics[width=\linewidth]{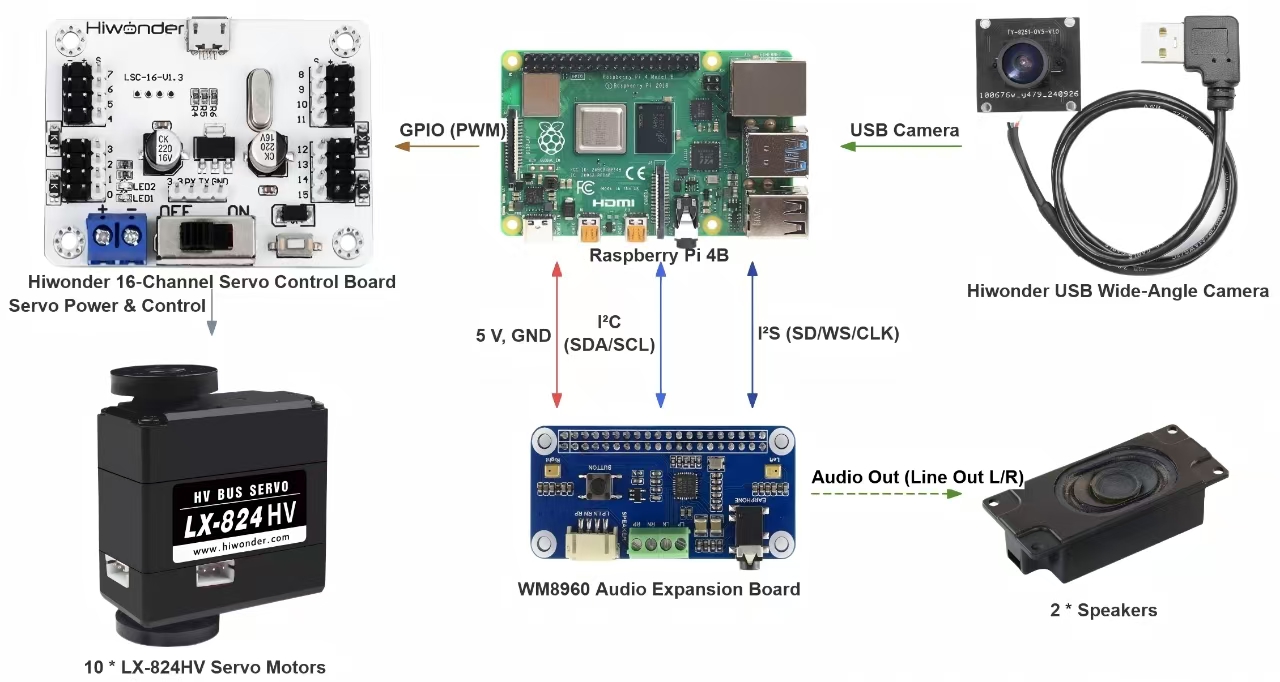}
    \caption{Hardware architecture and wiring layout of the embodied robot tutor.}
    \label{fig:hardware_structure}
\end{figure}

The robot's limbs are comprised of 10 LX-824HV servo motors connected to a PCA9685 PWM controller, communicating with the Raspberry Pi core board via GPIO. The expressive upper limb movements required for gesture interaction, such as waving, nodding, or raising the hand, have been pre-designed with servo motor IDs and angles for each servo motor, and the limbs return to their original positions after completing the movements. Each motor has a predefined motion sequence representing specific gestures such as excitement, engagement, or sadness. When the Raspberry Pi receives the gesture instructions from the server hosting the LLaMA model, it invokes the corresponding action packages through the action group controller which contains a PWM angle sequence, transmits it via the GPIO interface to the PCA9685 control board, and then to the servo motors, thereby executing the corresponding robot movements (see Figure~\ref{fig:hardware_structure}).

\begin{figure}[H]
    \centering
    \includegraphics[width=\linewidth]{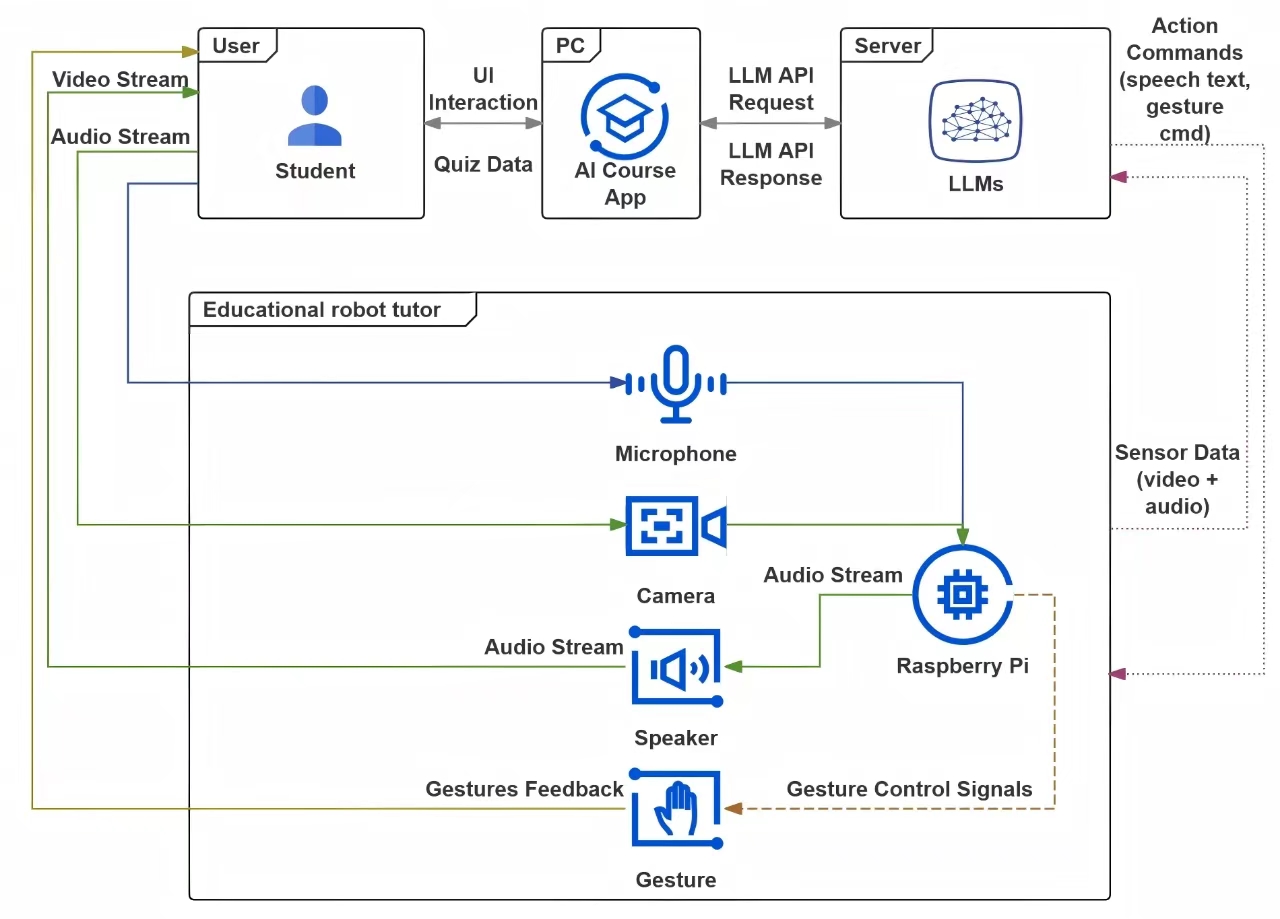}
    \caption{Data flow between the student, AI course app, LLM server, and educational robot tutor.}
    \label{fig:flowchart}
\end{figure}

The robot agent and laptop communicate with each other using HTTP requests over the same WIFI network. The flow of data is as follows and is shown in the chart  (see Figure~\ref{fig:flowchart}).


The student should say “Hi Rick” (The name of the robot tutor to enforce anthropomorphism) to start the conversation with the robot then formally begin the course. After receiving the greeting from the student, the robot tutor will reply to the greeting with a verbal introduction of itself then coming the course structure and the test summary to eliminate any novelty associated with the robot and prepare students for the experiment. The script of this introduction was as follows:

\textit{“Hello. My name is Rick. We will be going through a series of slides of the famous western story called "The Apology of Socrates" together with a brief test containing 5 questions on the screen. I appreciate you taking the time out of your schedule to work with me. Get ready so we can start. Tell me to start when you are ready.”} 

Students then navigate through the course slides which articulate the content regarding the story of The Apology of Socrates. After the slides end, the student will have the chance to ask a few questions to the robot tutor and get answers back in a human-like empathetic manner (the level of empathy depends on the trial explained later). Finally, students will be guided to the quiz questions, which consist of multiple-choice history questions of varying aspects of the story. As each student progresses through the test, their interaction with the laptop was communicated to the LLM server and the robot via an HTTP request. The communication protocol between the laptop (containing the servers) and the robot tutor allows for real time performance monitoring. Therefore, the robot can receive the instruction of the verbal content and gestures to be conveyed to the student with personalization. The answers to the questions will be checked by the course app on the server to determine if they are correct or not. The results will then be transmitted from the app server to the LLM server to generate empathic response content for the robot tutor. Then the response will be sent to the Raspberry Pi board which controls the robot’s behavior, giving positive reinforcement if the student answers the question correctly. After completing the test, a message is sent to the robot, at which point it shows its gratitude and gives a farewell. 

\begin{figure*}[t]
    \centering
    \includegraphics[width=\textwidth]{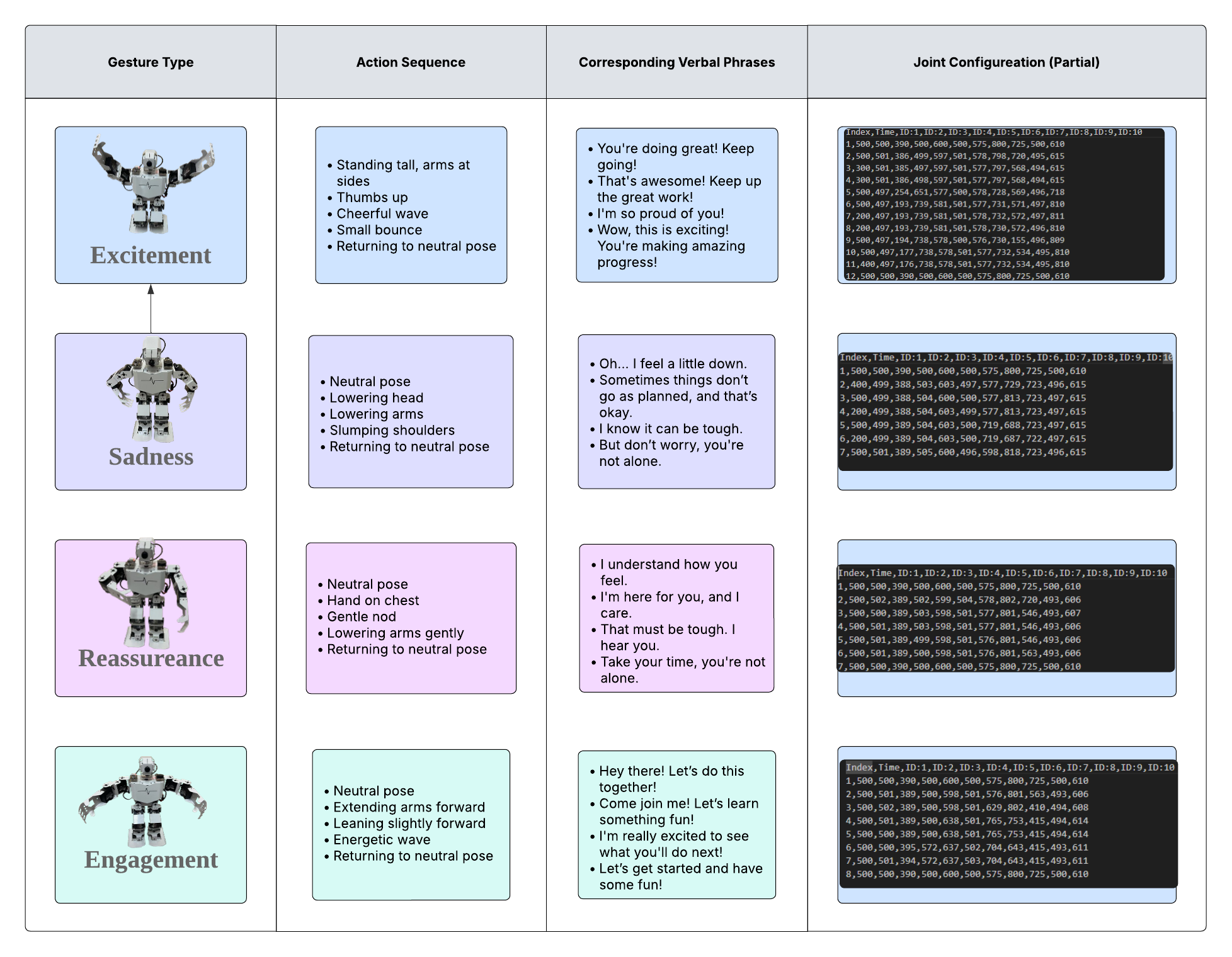}
    \caption{Sample gestures and corresponding verbal content used by the robot tutor to interact with students. Each pair is designed to enhance clarity, emotional resonance, and contextual appropriateness during the lesson delivery.}
    \label{fig:gestures}
\end{figure*}

In this study, we aim to enhance student engagement level and improve learning outcomes by incrementally integrating human-like empathetic capabilities into the robot tutor. These capabilities include synchronized verbal and non-verbal behaviours tailored to the learner's profile and real-time interaction context. The robot extends both arms forward, lean slightly to show interest and moves arms side to side in an energetic wave to engage with the students or slightly lower its head with slow gentle slump shoulder to express sadness when talking about the death-penalty verdict of Socrates. During the Q\&A section and the quiz, the robot can show positive reinforcement like excitement by raising one hand in a “thumbs up” position and raise another arm for a cheerful wave then bounce slightly if the student answers the question successfully. To the contrary, the robot tutor can gently nod, pair with hand placed near the chest and softly lower arms to the side to express understanding for the wrong answer. Figure~\ref{fig:gestures} illustrates the robot’s sample gestures and corresponding verbal contents in various settings. The engagement level is measured based on three different perspectives  (see Table~\ref{tab:engagement_types}), cognitive engagement, emotional engagement and behavioral engagement. Cognitive engagement focuses on the degree to which students maintain attention and effectively process information during interaction. Emotional engagement refers to the emotional responses to the conversation partner and behavioral engagement observes the non-verbal behavior. Facial expression, frequency of actions and eye movement will be monitored and video recorded. After the course, quizzes together with a questionnaire (see Appendix~\ref{appendix:questionnaire}) are delivered to the student to also measure their engagement level and compare learning outcomes.

\begin{table}[H]
\centering
\caption{Types of Engagement and Their Measurement}
\label{tab:engagement_types}
\resizebox{\columnwidth}{!}{%
\begin{tabular}{@{}>{\raggedright\arraybackslash}p{2.2cm}
                >{\raggedright\arraybackslash}p{3.2cm}
                >{\raggedright\arraybackslash}p{3.2cm}
                >{\raggedright\arraybackslash}p{3.2cm}@{}}
\toprule
\textbf{Engagement Type} & \textbf{Indicators} & \textbf{Measurement Tools} & \textbf{Data Type} \\
\midrule
Cognitive  & Focus, Attention, Information Processing & Eye Tracking, Cognitive Tests & Quantitative (Eye movement, Task performance) \\
\addlinespace
Emotional  & Interest, Enjoyment, Frustration & Facial Recognition Software, Self-Report Scales & Quantitative \& Qualitative (Facial expressions, Survey responses) \\
\addlinespace
Behavioral & Verbal Responses, Gestures, Activeness & Interaction Logs, Video Analysis & Quantitative \& Qualitative (Frequency of actions, Video coding) \\
\bottomrule
\end{tabular}
}
\end{table}

The robot tutor dynamically adjusts its verbal and non-verbal behaviours in response to the student's current emotional and cognitive state, as well as contextual factors within the learning interaction. These adaptive responses are informed by each student's learning profile, which captures prior engagement data and interaction history. For the experimental design, we utilize the same test and environment setup across all students (so that the course materials and quiz questions are all the same). The only thing that changes among trials is the level of empathetic capability. By progressively increasing the empathetic capabilities of the robot tutor, this experiment seeks to analyze the impact of integrating the memory architecture (for personalization), emotional intelligence and gestures on student engagement and learning outcomes.

\section{Research Questions and Trials}
We developed the following guiding research questions to explore the above phenomenon:

\begin{itemize}
    \item \textbf{RQ1:} Does the integration of verbal communication with non-verbal gestural behaviors improve student engagement level compared to verbal-only interactions?
    \item \textbf{RQ2:} Can memory-based personalization further enhance engagement level by adapting the robot's behaviors with reference to student profiles? (In addition, we will explore whether memory-based personalization or gestural interaction character implemented on the robot offers greater benefits to student engagement.)
    \item \textbf{RQ3:} Do these increasing levels of empathetic interaction translate into measurable improvements in learning process \& outcomes with the help of our Engagement Vector?
\end{itemize}

Students interacted with the robot tutor under three experimental conditions, each introducing progressively more empathetic features:

\begin{itemize}
    \item \textbf{Trial 1:} The robot tutor teaches using verbal communication only, without any gestural cues and memory-driven personalization.
    \item \textbf{Trial 2:} The robot tutor is designed to deliver the class using both verbal and non-verbal (gestural) communication. In an additional test, we implement a personalized verbal interaction mode based on memory, without gestures and compare its effects against the previous condition, which combined gesture-enabled but non-personalized verbal communication.
    \item \textbf{Trial 3:} The robot tutor employs verbal and gestural capabilities while using memory to personalize interactions based on the student’s profile to better convey empathy. 
\end{itemize}

\section{Methodology and Analysis Framework}

\begin{figure*}[t]
    \centering
    \includegraphics[width=\textwidth]{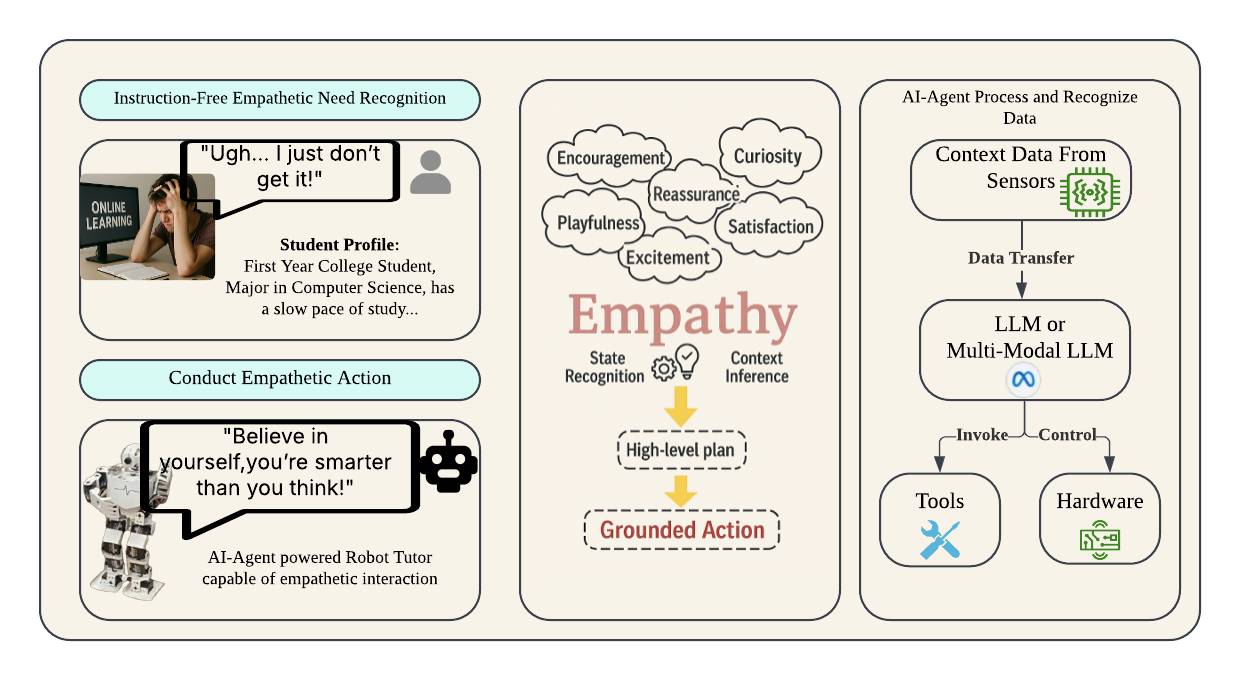}
    \caption{Conceptual framework illustrating how the AI-Agent-powered robot tutor recognizes student needs through contextual sensing and empathetic inference. The system integrates multimodal LLMs, student profiling, and emotion recognition to plan and deliver grounded empathetic actions, enabling natural and supportive educational interactions.}
    \label{fig:teaching_scenario}
\end{figure*}

To evaluate the effectiveness of an educational robot tutor in mimicking human emotion, memory and gestures in an observed scenario to enhance the interaction quality with students then achieving better outcomes, an experimental study was conducted with the help of an AI-Agent (see Figure~\ref{fig:teaching_scenario}). The experiment aims to investigate whether the presence of a human-like empathetic robot tutor would increase student’s engagement level and foster the learning outcome during a study task. A three-trial experiment design was employed. To ensure that the knowledge is evenly distributed among the groups, the subjects are selected at random. A total of 15 college students took part in Trial 1 of the experiment; this includes both females and males aged from 18-25 years old (m= 22 \(\sigma = 2.4 \), male: 7, female: 8). Another 15 college students took part in Trail 2 of the experiment; this consisted of both females and males in the age range of 19-25 years old (m= 23.7, \(\sigma = 2.3\), male: 6, female: 9). A final group of 15 students took part in Trial 3 with females and males aged from 18-26 (m= 24.1, \(\sigma = 2.1\), male: 8, female: 7). The additional test group of the same number of students (m= 24.8, \(\sigma = 3.4\), male: 9, female: 6).

\subsection{Data Collection and Analysis}
To quantitatively evaluate engagement across the experiments, we implement an Engagement Vector Model. Each aspect of engagement, including cognitive (Equation (2)), emotional (Equation (3\&4)), and behavioural (Equation (5)), is normalized to a range from 0.0 (low level) to 1.0 (high
level) based on student’s performance data collected in the trials. For producing a single, interpretable engagement score, a weighted average of the three dimensions was calculated (Equation (6)). 

\paragraph{Engagement Vector}
Each student's engagement level at a given trial $n$ is represented by a 3-dimensional vector:
\begin{equation}
\mathbf{E}^{(n)} = 
\begin{bmatrix}
E_\text{cog}^{(n)} \\
E_\text{emo}^{(n)} \\
E_\text{beh}^{(n)}
\end{bmatrix}
\in [0,1]^3
\end{equation}
where $E_\text{cog}^{(n)}$, $E_\text{emo}^{(n)}$, and $E_\text{beh}^{(n)}$ denote the \textbf{Cognitive}, \textbf{Emotional}, and \textbf{Behavioral} engagement scores respectively, normalized to the range $[0,1]$.

\paragraph{Normalization of Engagement Metrics}

Raw engagement indicators are linearly transformed to the $[0,1]$ range using direct scaling.

\subparagraph{Cognitive Engagement:}
\begin{equation}
E_\text{cog} = \lambda_1 \cdot \left(1 - \frac{T_q - T_{\min}}{T_{\max} - T_{\min}}\right)
+ \lambda_2 \cdot \left(\frac{S_q}{100}\right)
+ \lambda_3 \cdot \left(\frac{G_f}{100}\right)
\end{equation}

\begin{itemize}
  \item $T_q$ = Quiz completion time (min), inverted to favor faster responses
  \item $S_q$ = Quiz success rate (\%)
  \item $G_f$ = Gaze fixation ratio (\%), calculated from Tobii Eye Tracker 5's movement map.
  \item $\lambda_i$ = weighting coefficients, $\sum \lambda_i = 1$, to ensure normalized contributions
\end{itemize}

\subparagraph{Emotional Engagement:}
Emotional engagement is calculated using a simplified model that fuses facial expression analysis and subjective feedback. Instead of treating positive and negative emotions separately, we define a net emotional valence score:

\begin{equation}
\tilde{V}_e = \frac{Pe - Fr + 100}{200}
\end{equation}

\begin{equation}
E_{emo} = \gamma_1 \cdot \tilde{V}_e + \gamma_2 \cdot \left( \frac{Rs - 1}{4} \right)
\quad \text{with} \quad \gamma_1 + \gamma_2 = 1
\end{equation}

\begin{itemize}
  \item $P_e$ = Percentage of positive expressions (e.g., smiling), use a pre-trained model called DeepFace in Python.
  \item $F_r$ = Percentage of frustrated expressions (e.g., angry, sad, disgust), also use DeepFace to detect
  \item $R_s$ = Self-report rating on a 1–5 scale
  \item $\gamma_i$ = weighting coefficients, $\sum \gamma_i = 1$
\end{itemize}

The valence score is then combined with the student’s normalized self-report rating \( Rs \in [1, 5] \) to compute the final emotional engagement score.

This fusion ensures that both observed (facial expression) and perceived (self-reported) emotional cues are reflected in the final emotional engagement score.

\subparagraph{Behavioral Engagement:}
\begin{equation}
E_\text{beh} = \beta_1 \cdot \left(\frac{I_f}{I_{\max}}\right)
+ \beta_2 \cdot \left(\frac{G_a}{100}\right)
+ \beta_3 \cdot \left(\frac{V_r}{100}\right)
\end{equation}

\begin{itemize}
  \item $I_f$ = Number of questions or spontaneous student queries to the robot recorded in the log of the application (normalized by $I_{\max}$, e.g., 5)
  \item $G_a$ = Gesture activity time ratio (\%) during the whole interaction process
  \item $V_r$ = Percentage of robot prompts that receive a student voice reply
  \item $\beta_i$ = weighting coefficients, $\sum \beta_i = 1$
\end{itemize}

\paragraph{Weighted Fusion}
The scalar \textit{Final Engagement Score} is computed using a weighted sum of the three components. Given our focus on evaluating the integrated effects of empathetic traits in robots, equal weights provide a balanced and reproducible framework. This avoids introducing additional model complexity and future research may consider adaptive or data-driven weighting strategies.

\begin{equation}
E_\text{final}^{(n)} = w_\text{cog} \cdot E_\text{cog}^{(n)} + 
w_\text{emo} \cdot E_\text{emo}^{(n)} + 
w_\text{beh} \cdot E_\text{beh}^{(n)}
\end{equation}

In our study, uniform weighting is assumed:
\[
w_\text{cog} = w_\text{emo} = w_\text{beh} = \frac{1}{3}
\]

Previous studies have explored the measurement of individual engagement components using various methods. For example, biometric sensors are used to assess emotional and behavioural engagement and machine learning algorithms are utilized to predict cognitive engagement based on existing data set. However, these methodologies often treat each aspect of the engagement separately and can not provide a holistic assessment of the HRI process.

The innovation of our proposed  \textbf{Engagement Vector Model } lies in its ability to synthesize multiple engagement indicators into a single, dynamic metric for educational settings. Together with normalizing diverse data streams and applying weighted coefficients, this model statistically facilitates the research to be more all-sided and immediate comprehensible.

\subsection{Engagement Analysis}
Engagement is analyzed across three aspects: cognitive, emotional and behavioral. Cognitive engagement was assessed using eye tracking data and task performance metrics (quiz completion time and success rate). Emotional engagement was evaluated using facial recognition software and self-reported scales which measured indicators such as interest, enjoyment. Behavioral engagement was measured through video analysis and interaction logs, which tracked gestures, and overall activeness.  These indicators provided insight into the student’s focus, attention, and emotional responses during the test. 
\begin{itemize}
    \item \textbf{Trial 1: Verbal Only Interaction} 
    
    \textit{Cognitive Engagement}: Eye-tracking data revealed frequent gaze shifts and prolonged fixations away from the screen and the robot tutor, indicating lower levels of focus and less sustained attention. This suggests that the students were not fully engaged within the course content. The quiz completion time averaged 8.3 minutes, which was the longest among all trials, showing that students required more time to process the information and complete the task. What is more, success rates in this trial were lower compared to the others, indicating that the lack of non-verbal cues and emotional engagement negatively impacted student’s cognitive processing (see Figure~\ref{tab:engagement_metrics}).
    
    \textit{Emotional Engagement}:  Facial recognition software revealed higher levels of frustration among students, with a noticeable lack of positive expressions, such as smiling or nodding. This suggests that the purely verbal interaction with the robot did not elicit much emotional connection. Self-reports on the engagement atmosphere fostered by the robot, on a scale from 0-1, averaged 0.4/1, indicating low enjoyment and a rigid emotional response. Students in this trial were easily bored and expressed dissatisfaction with the robot tutor’s teaching style, since the verbal- only approach failed to establish a connection beyond simply transferring information.

    \textit{Behavioural Engagement}:  During this trial, where the robot only used verbal cues, students initiated an average of 8 interactions during the Q\&A session. The interaction logs showed relatively low engagement. Body movement analysis from video recordings revealed low activeness, since students exhibiting minimal physical responses (e.g., little to no gesturing or leaning forward). This suggests that the verbal-only interaction failed to significantly stimulate active participation or maintain student’s attention throughout the task

\begin{figure*}[t]
\centering
\caption{Engagement and Performance Metrics Across Trials}
\label{tab:engagement_metrics}
\resizebox{1.0\textwidth}{!}{%
\begin{tabular}{|l|c|c|c|}
\hline
\textbf{Metric} & \textbf{Trial 1 (Verbal Only)} & \textbf{Trial 2 (Verbal + Gestural)} & \textbf{Trial 3 (Verbal + Gestural + Memory)} \\
\hline
\textbf{AVG Quiz Completion Time (min)} & 8.3 & 7.5 & 6.3 \\
\hline
\textbf{Quiz Success Rate (\%)} & 50\% & 66\% & 78\% \\
\hline
\textbf{Emotional Engagement (0--1)} & 0.4 & 0.6 & 0.75 \\
\hline
\textbf{Behavioral Engagement (interactions)} & 8 & 9 & 11 \\
\hline
\textbf{Satisfaction Score (0--1)} & 0.3 & 0.6 & 0.75 \\
\hline
\end{tabular}%
}
\end{figure*}

     \item \textbf{Trial 2: Verbal + Gestural Interaction}  

     \textit{Cognitive Engagement}: With the addition of non-verbal cues (gestures), eye-tracking data showed an improvement in sustained focus. Students exhibited fewer distractions and spent more time focusing on the screen and the robot, suggesting a positive shift in cognitive engagement. The quiz completion time was reduced to 7.5 minutes, indicating better performance and quicker processing of the test material. Success rates also improved, reflecting the added benefits of gestures assisted communication in maintaining student’s attention.  This suggests that non-verbal communication can help guide student’s focus and enhance the cognitive processing of the material (see Figure~\ref{tab:engagement_metrics}). 

     \textit{Emotional Engagement}: With the addition of gestural cues, facial recognition indicated a decrease in frustration levels, and positive emotional expression such as smiling and eye gaze increased slightly. This enhancement in emotional engagement suggested that the gestures helped create a more empathetic and dynamic interaction with the robot. Students' self-perceived engagement score increased to an average of 0.6/1, indicating moderate emotional engagement. However, some students still felt neutral or slightly positive, the combination of verbal and non-verbal cues helped them feel more involved and connected to the robot during the learning process based on the comment from post-test questionnaire.

     \textit{Behavioural Engagement}: In the trial where the robot combines both verbal and non-verbal cues, interaction frequency increased to 9 interactions per session, indicating that the students were more engaged than that in Trial 1. Video analysis revealed more dynamic body language, including leaning forward, nodding, and occasional gestures (such as hand movements). These kinds of behaviors suggest that the addition of gestures helped stimulate greater interaction and made students feel more involved in the learning process, as they responded to the robot frequently and with more physical expressiveness.

     \item \textbf{Trial 3: Verbal + Gestural + Memory-Based Personalization} 
     
      \textit{Cognitive Engagement}: In this trial, where the robot incorporated both verbal and non-verbal communication together with personalized memory-based interactions, eye-tracking data indicated the highest sustained attention, with minimal off-screen glances. This reflects that the students were deeply focused on the course and were fully immersed in the learning experience. The quiz completion time was the shortest across all groups, averaging 6.3 minutes, suggesting that students were able to handle quiz questions more quickly. In addition, success rates were the highest in this trial, supporting the idea that the robot’s capability to showcase personalized interactions and reflect on student’s preferences on the profile significantly boosted cognitive engagement and task performance (see Figure~\ref{tab:engagement_metrics}).

      \textit{Emotion Engagement}: In this trial, the robot utilized verbal and non-verbal gestural communications along with memory-based personalized interactions. Facial recognition showed a significant increase in positive expressions, including smiling, nodding, and sustained eye contact. Students appeared much more engaged and connected with the robot, demonstrating a higher emotional investment. Frustration levels were the lowest across all three trials, reflecting the robot’s ability to effectively adapt to individual student’s needs. Self-reported average engagement level rise to 0.75/1, indicating high emotional engagement. The students reported feeling more supported and motivated due to the robot’s personalized, empathetic responses, which enhanced the overall learning experience.

      \textit{Behavioural Engagement}: The highest level of behavioral engagement occurred in Trial 3, where the robot incorporated both verbal and gestural communication along with memory-based personalized interactions. In this trial, students engaged with the robot 11 times per session on average, showing a significant increase in interaction frequency. Video analysis highlighted markedly higher activeness, with students displaying a wide range of gestures, such as nodding frequently, smiling, hand movement, and leaning in. Additionally, students gave more verbal responses and actively responded to personalized feedback during the Q\&A session, indicating a higher level of behavioral engagement in the learning process. 

\end{itemize}

\subsection{Student Perception}
Student perceptions of the learning experience were evaluated through a post-experiment questionnaire, measuring overall satisfaction, engagement, and perceived effectiveness of the robot tutor. The results highlight how students responded to different levels of empathetic capabilities in the robot. 

Trial 1: Students reported an average satisfaction score of 0.3/1, the lowest among the three trials. Feedback indicated that the robot was perceived as “monotonous” and “uninspiring”, with many students expressing frustration over its lack of responsiveness and empathetic connection. Many participants commented that the interaction felt one-directional instead of bi-directional, as the robot simply delivered information without catering for student's needs, leading to lower motivation and engagement during the session. 

Trial 2's satisfaction levels improved to 0.6/1, showing moderate engagement. Students appreciated the usage of gestures, which made interactions feel more nature and engaging compared to Trial 1. However, some participants still complained that the robot lacked human-like traits compared to human tutor, stating that while gestures helped sustain attention, the interaction remained somewhat generic. Several students mentioned that a more adaptive response system could further enhance engagement. 

Students from Trial 3 reported the highest satisfaction with an average score of 0.75/1, indicating a strong sense of support and engagement (see Figure~\ref{tab:engagement_metrics}). Many participants expressed that the personalized interactions made them feel valued and understood, as the robot remembered past responses and students’ preferences to adapt feedback accordingly. Comments highlighted how dynamic and responsive the robot felt in this trial, with students stating that the tailored encouragement made the learning experience more enjoyable and effective. A few students said that while the interactions were highly engaging, balancing personalization with instructional efficiency could be further optimized.

\subsection{Comparison of the vectors throughout trails}

\begin{figure}[h]
    \centering
    \includegraphics[width=\linewidth]{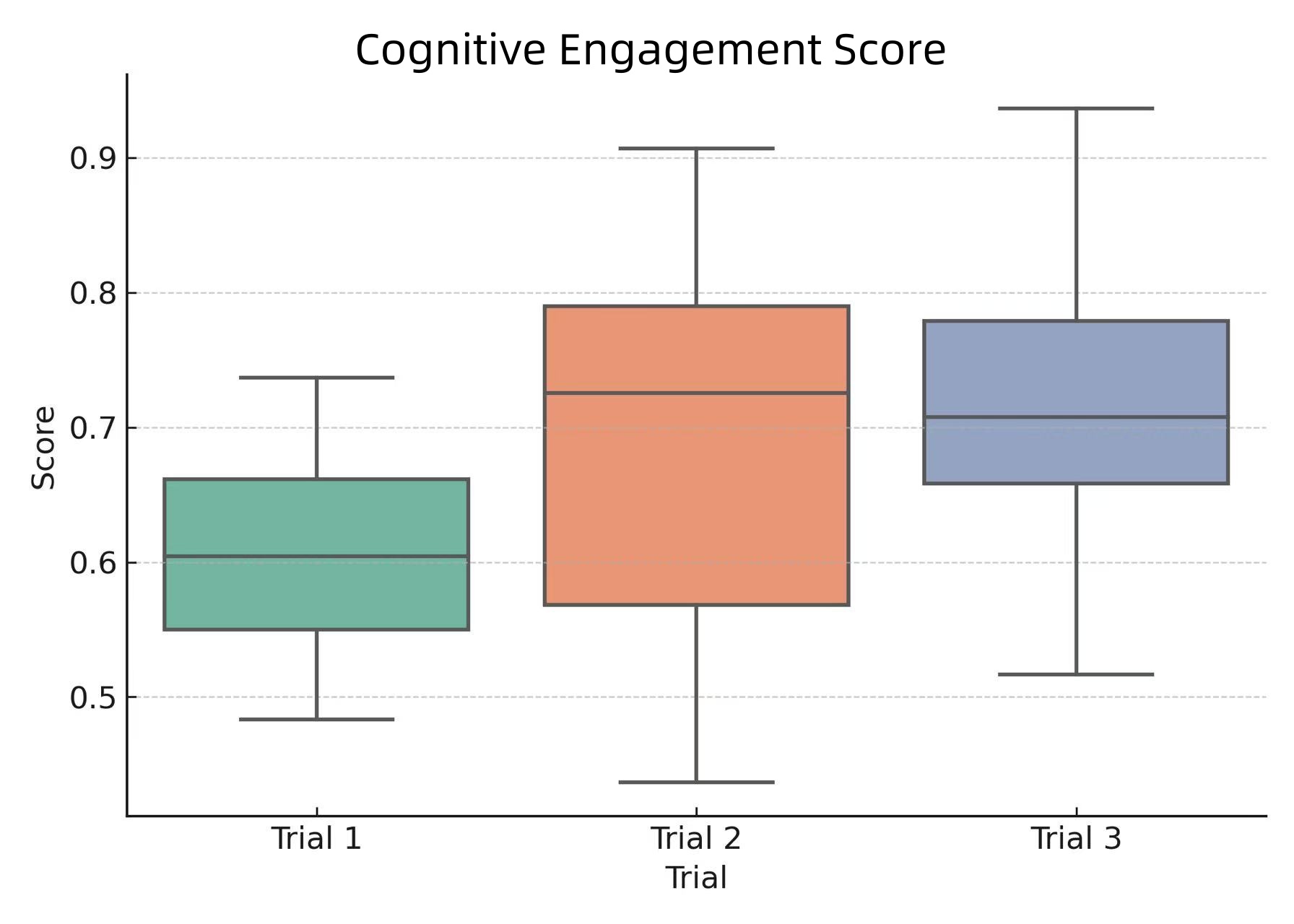}
    \caption{Distribution of Cognitive Engagement Scores across the three trial groups.}
    \label{fig:cognitive_boxplot}
\end{figure}

To better understand cognitive engagement across the three trials, we visualized the group scores using a boxplot with reference to formula (2), as shown in Figure~\ref{fig:cognitive_boxplot}.

\noindent
From the figure, we observe that:
\begin{itemize}
    \item \textbf{Trail 1} shows lower overall engagement, with both median and upper whisker values below the other groups. 
    \item \textbf{Trail 2} has the highest median among all groups, but a larger spread, suggesting varied engagement levels among participants.
    \item \textbf{Trail 3} exhibits a relatively high median score and tight interquartile range, indicating consistent cognitive involvement.
\end{itemize}

These differences reflect how instructional design, such as multi-modal interaction and adaptive feedback, may have influenced students’ depth of cognitive processing.

\begin{figure}[h]
    \centering
    \includegraphics[width=\linewidth]{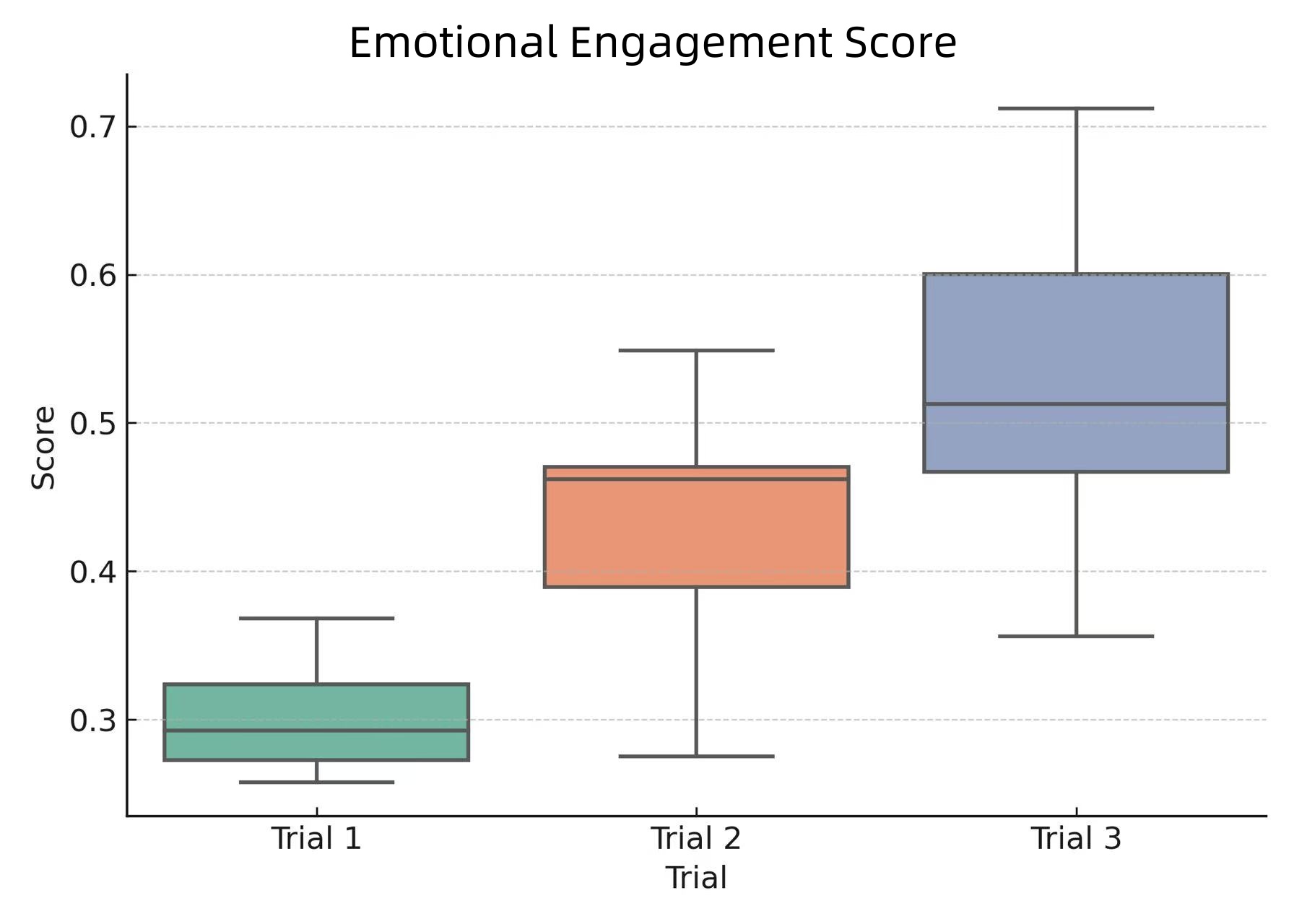}
    \caption{Distribution of Emotional Engagement Scores across the three trial groups.}
    \label{fig:emotional_boxplot}
\end{figure}

Figure~\ref{fig:emotional_boxplot} displays the distribution of emotional engagement scores across the three experimental groups.

\noindent
From the plot, we can observe the following trends:
\begin{itemize}
    \item \textbf{Trail 1} records the lowest emotional engagement scores, both in terms of central tendency and spread.
    \item \textbf{Trail 2} shows moderately lower engagement than Trial 3, with slightly reduced variability.
    \item \textbf{Trail 3} demonstrates the highest overall emotional engagement, with both a high median score and broader interquartile range.
 
\end{itemize}

These results suggest that participants in Trial 3 experienced more positive affect and emotional resonance during the session. In contrast, Trial 1 \& 2 might have lacked sufficient emotional stimulation or perceived relevance. The scoring model integrates both externally detected emotion and internal feedback, reinforcing the robustness of the observed trend.

Behavioral engagement refers to the extent to which learners physically and socially participate in the learning task, such as through gesture use, interaction frequency, and verbal responses. As defined in Equation~(5), the behavioral engagement score $E_{\text{beh}}$ is calculated based on three main factors: interaction frequency ($I_f$), gesture activity ratio ($Ga$), and verbal response rate ($Vr$).

\vspace{1em}

\begin{figure}[h]
    \centering
    \includegraphics[width=0.92\linewidth]{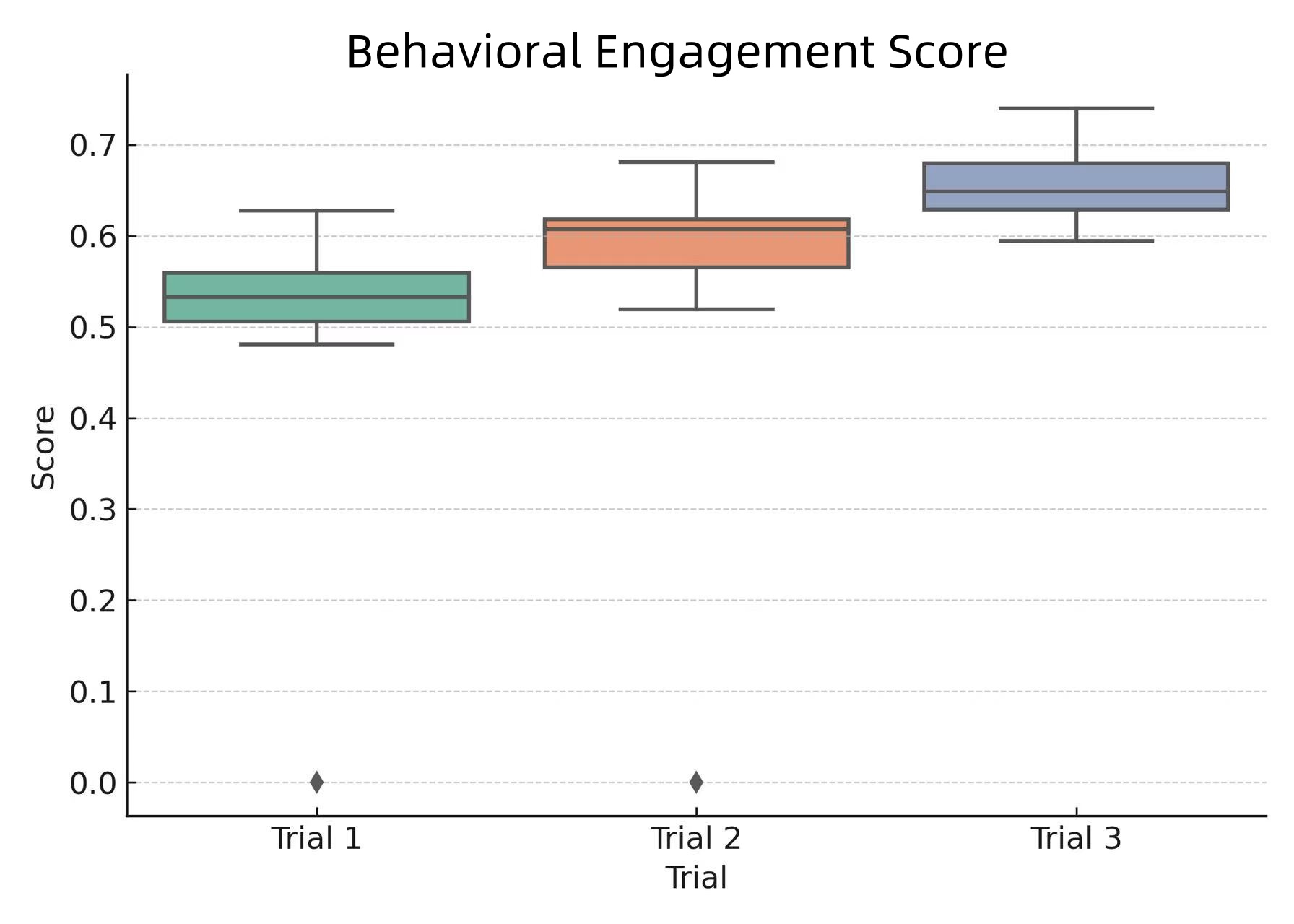}
    \caption{Distribution of Behavioral Engagement Scores across the three trial groups.}
    \label{fig:behavioral_boxplot}
\end{figure}

To compare behavioral engagement across experimental conditions, we present a boxplot visualization of the engagement scores in Figure~\ref{fig:behavioral_boxplot}.

\noindent
As observed in the figure:
\begin{itemize}
    \item \textbf{Trail 1} (verbal-only interaction) displayed relatively constrained behavioral engagement, with limited variation in student interaction frequency and gesture response.
    \item \textbf{Trail 2}, which incorporated verbal and gestural communication, showed a notable increase in both median score and engagement consistency, suggesting that gesture integration stimulated physical participation.
    \item \textbf{Trail 3} demonstrated the highest behavioral engagement, with frequent and spontaneous interaction. The robot’s personalized prompts and memory-based adaptation encouraged students to respond more often, lean forward, and express themselves more openly.
\end{itemize}

These results highlight the effectiveness of integrating non-verbal cues and personalized interaction history in stimulating active learning behaviors. As behavioral engagement is crucial for maintaining focus and sustaining interest, these findings suggest that adaptive, empathetic designs can significantly elevate students' willingness to participate during robot-facilitated instruction.

\begin{figure}[h]
    \centering
    \includegraphics[width=\linewidth]{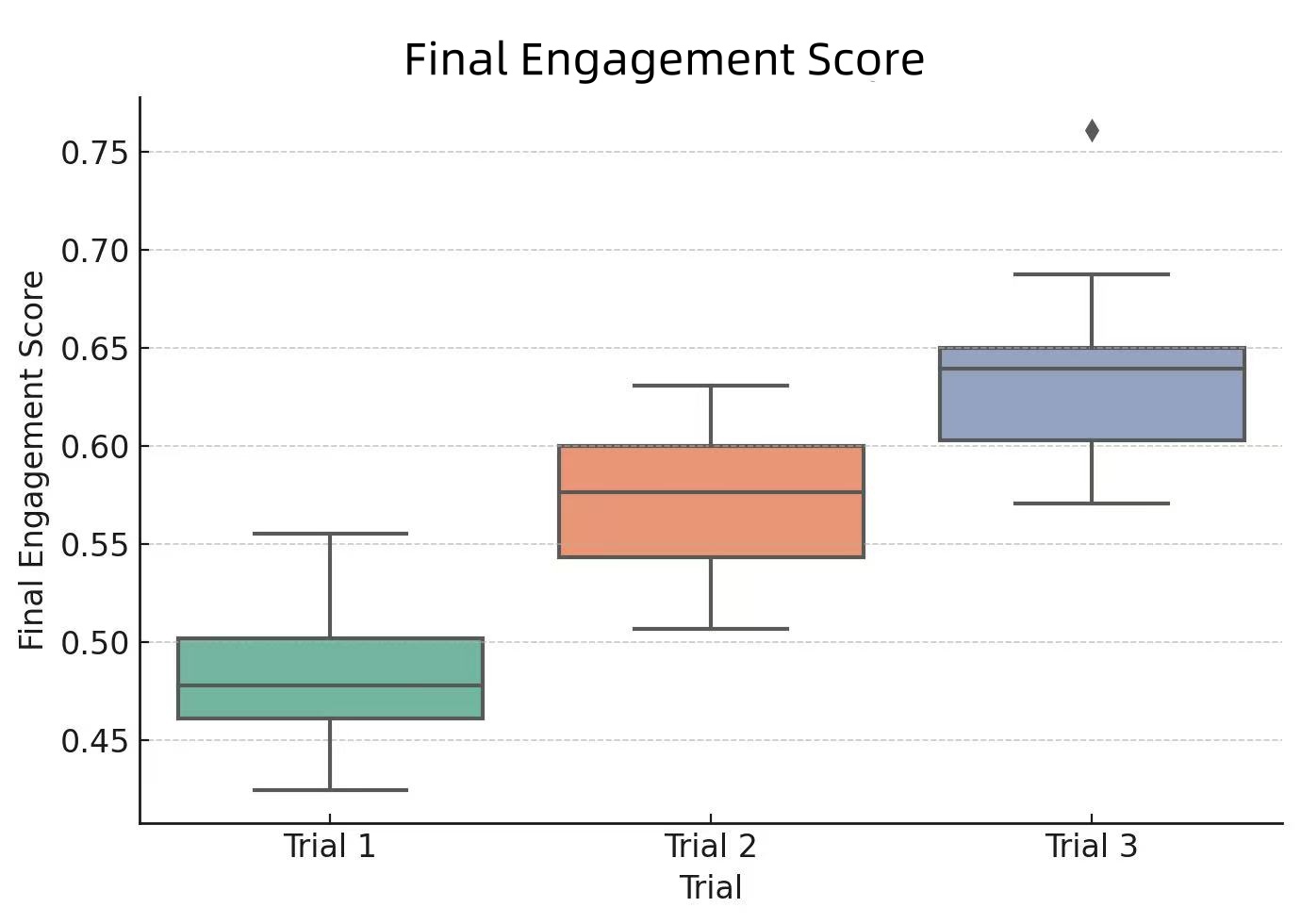}
    \caption{Final Engagement Score distribution across Trial 1 (verbal only), Trial 2 (verbal + gesture), and Trial 3 (verbal + gesture + memory).}
    \label{fig:final_engagement_boxplot}
\end{figure}

\begin{figure}[h]
    \centering
    \includegraphics[width=1.0\linewidth]{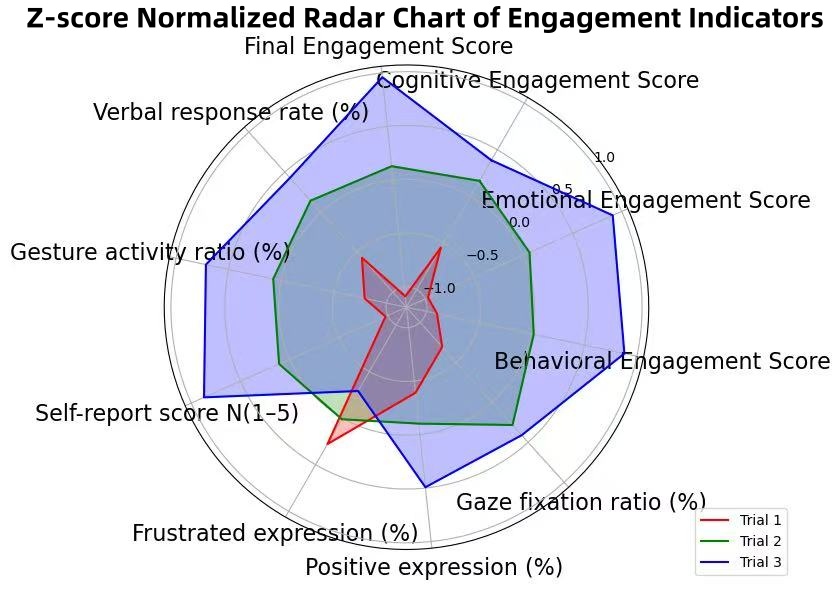}
    \caption{Z-score normalized radar chart illustrating engagement indicators across the three trial conditions. Each axis represents a key metric from cognitive, emotional, or behavioral engagement dimensions}
    \label{fig:radar}
\end{figure}

Figure~\ref{fig:final_engagement_boxplot} illustrates the distribution of final engagement scores across the three experimental conditions. The boxplot reveals a clear upward trend from Trial 1 to Trial 3, validating the effectiveness of incorporating human-like traits in robotic instruction design. The Z-score normalized radar chart (Figure~\ref{fig:radar}) also supports our statistical results and highlights the cumulative effect of empathetic features in enhancing student engagement, confirming the effectiveness of multi-modal and memory-driven HRI strategies.

\noindent
Key insights from the comparison include:
\begin{itemize}
    \item \textbf{Trial 1}, where the robot used only verbal instruction, showed the lowest engagement scores and limited variation.
    \item \textbf{Trial 2} achieved moderate improvements, benefiting from the addition of non-verbal cues such as gestures.
    \item \textbf{Trial 3} significantly outperformed the other groups, with the highest median and reduced variance—indicating both stronger and more consistent engagement due to personalized memory-based interaction.
\end{itemize}

These results provide concrete evidence that integrating multiple human-like traits—particularly memory-driven personalization and non-verbal expressiveness—greatly enhances student engagement. The fusion score model enables a comprehensive understanding of the robot’s pedagogical impact beyond individual dimensions.

\subsubsection{Trial-Wise Statistical Comparison}

To determine whether engagement levels significantly differed across the three trial conditions, we performed pairwise Mann-Whitney U tests (MWU) for each engagement metric. The results, summarized in Table~\ref{tab:mwu_trimmed}, show both descriptive statistics and statistical comparisons between trials.

\begin{table}[t]
\centering
\scriptsize
\caption{MWU Test Results Between Trials (p-values)}
\label{tab:mwu_trimmed}
\resizebox{\columnwidth}{!}{%
\begin{tabular}{@{}llccc@{}}
\toprule
\textbf{Metric} & \textbf{Trial} & \textbf{vs T2 (p)} & \textbf{vs T1 (p)} & \textbf{vs T3 (p)} \\
\midrule
Cognitive  & T1 & 0.0679   &         & 0.00617   \\
Emotional  & T1 & 0.000189 &         & 6.15e-06  \\
Behavioral & T1 & 0.00479  &         & 1.94e-05  \\
Cognitive  & T2 &          & 0.0679  & 0.678     \\
Emotional  & T2 &          & 0.000189& 0.00952   \\
Behavioral & T2 &          & 0.00479 & 0.00174   \\
Cognitive  & T3 & 0.678    & 0.00617 &           \\
Emotional  & T3 & 0.00952  & 6.15e-06&           \\
Behavioral & T3 & 0.00174  & 1.94e-05&           \\
\bottomrule
\end{tabular}
}
\end{table}

\noindent
From the results above, we observe that:
\begin{itemize}
    \item \textbf{Cognitive Engagement} showed a statistically significant difference only between Trial 3 and Trial 1 ($p = 0.0062$), but not between other pairs.
    \item \textbf{Emotional Engagement} was significantly different across all trial pairs, with Trial 3 consistently outperforming Trials 2 and 1.
    \item \textbf{Behavioral Engagement} differences were also significant across all trials, with Trial 3 showing the highest scores, indicating stronger physical and interactional involvement.
\end{itemize}

\begin{table}
\renewcommand{\arraystretch}{3.5} 
\centering
\caption{Comparison Between Gestural and Memory-Based Personalization Conditions}
\label{tab:gesture_memory}
\resizebox{0.5\textwidth}{!}{%
\begin{tabular}{|l|c|c|}
\hline
\textbf{{\LARGE AVG Score}} & \textbf{{\LARGE Gesture Enabled}} & \textbf{{\LARGE Personalization Enabled}} \\
\hline
\textbf{{\LARGE Cognitive Engagement}} & {\LARGE 0.69} & {\LARGE 0.75} \\
\hline
\textbf{{\LARGE Emotional Engagement}} & {\LARGE 0.43} & {\LARGE 0.41} \\
\hline
\textbf{{\LARGE Behavioral Engagement}} & {\LARGE 0.61} & {\LARGE 0.50} \\
\hline

\end{tabular}%
}
\end{table}

Besides the core trials, we conducted a focused comparison to determine whether gestural or memory-driven personalization interaction contributes more to student engagement and learning outcome. The final results (see Table~\ref{tab:gesture_memory}) suggest a nuanced trade-off between these modalities. The personalized verbal interaction condition led to a slightly higher cognitive engagement score (0.75 vs 0.69) and comparable emotional engagement, indicating that customizing the robot's responses to individual student profile can help to concentrate more effectively and process content with greater depth. However, the gesture enhanced condition results in stronger behavioral engagement (0.61 vs 0.50). This implies that visible non-verbal cues play an essential role in making interactions feel more expressive and interactive, eliciting physical responsiveness.

These results provide strong statistical evidence that enhanced robot features—such as gestures and personalized memory—used in later trials meaningfully elevated student engagement.

\section{Conclusion}

All three research questions are strongly supported by our findings. For RQ1, involving gestures in Trial 2 led to improved focus, interaction, and overall engagement scores (0.58 vs 0.48) compared to verbal-only interaction, although memory-driven personalization is still missing. For RQ2, personalization in Trial 3 further boosted engagement level (0.64) and satisfaction (0.75), showcasing the benefits of adaptive, student-aware responses. Addressing RQ3, we observed clear gains in learning process \& outcomes across trials, higher quiz correct rates and shorter completion times which demonstrates that empathetic interaction correlates with improved performance, as captured by our Engagement Vector Model. The additional experiment showed that memory-driven personalization slightly improved cognitive engagement, while gestural interaction elicited better behavioral engagement, highlighting a trade-off between personalized depth and expressive interactivity.

Overall, the study indicates that a multimodal, adaptive robotic tutor equipped with human-like traits can serve well to enhance students' learning outcomes by creating a more interactive, responsive, and personalized educational environment. What is more, the Engagement Vector Model that captures interaction data comprehensively provides an efficient way to judge the quality of HRI from the manipulation of multi-dimension data then shed light on how to measure and optimize the performance of robot in educational scenarios. Future research should focus on improving the degree of personalization and exploring how to implement human-like traits in ways that support, rather than overwhelm, students which was something that was found to detract from the benefits of the robot interaction in our study. Additionally, integrating a broader range of human-like traits to further personalize robot tutors presents a promising direction for continued development.

\bibliography{references}

\clearpage 
\appendix
\section{Appendix: Course App Interface}
\label{appendix:courseapp}

\begin{figure}[H]
    \centering
    \includegraphics[width=0.6\textwidth]{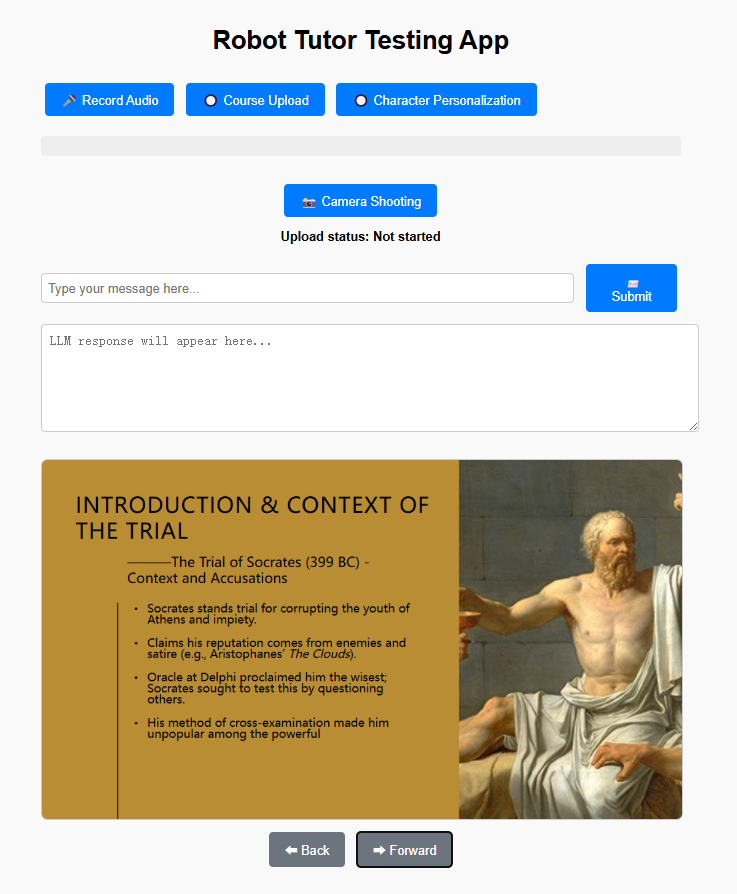}
    \caption{
    User interface of the Robot Tutor Course App. 
    The \textbf{Record Audio} button enables the robot to start listening to the student's speech in real time. 
    The \textbf{Course Upload} button allows scaling the robot tutor to cover multiple subjects. 
    The \textbf{Character Personalization} button adapts the robot's persona to suit individual student needs. 
    The \textbf{Camera Shooting} button activates facial expression capture for inferring the student's emotional state. 
    The text input box provides an alternative input method when speech recognition fails, due to strong accents. 
    The \textbf{LLM response} section displays the robot's textual reply as a complement to its spoken response, ensuring clarity for the student. 
    The course slide at the bottom illustrates a history lesson on the Trial of Socrates, demonstrating the system's educational content delivery.
    }
    \label{fig:course_app_ui}
\end{figure}

Figure~\ref{fig:course_app_ui} illustrates the user interface of the Course App used during the robot tutoring session. 

\clearpage

\section{Appendix: Post-Experiment Questionnaire}
\label{appendix:questionnaire}

\textbf{Section 1: Engagement} \\
Q1. The robot tutor maintained my attention throughout the session. \hfill ( ) \\
Q2. The robot’s gestures or expressions helped me stay engaged with the content. \hfill ( ) \\

\textit{Scale: 1 = Strongly Disagree to 5 = Strongly Agree}

\vspace{1em}
\textbf{Section 2: Satisfaction} \\
Q3. I enjoyed learning with the robot tutor. \hfill ( ) \\
Q4. I would be interested in using this kind of robot tutor in other classes. \hfill ( ) \\

\textit{Scale: 1 = Strongly Disagree to 5 = Strongly Agree}

\vspace{1em}
\textbf{Section 3: Perceived Effectiveness} \\
Q5. The robot adapted its behavior based on my responses or emotional state. \hfill ( ) \\
Q6. I understood the topic better after the session with the robot tutor. \hfill ( ) \\

\textit{Scale: 1 = Strongly Disagree to 5 = Strongly Agree}

\vspace{1em}
\textbf{Section 4: Open-Ended Feedback} \\
Q7. What was the most effective or memorable aspect of the robot tutor? \\

Q8. What would you change or improve about the robot's behavior or teaching style? \\

\end{document}